\documentclass[11pt,a4paper]{article}
\usepackage[usenames,dvipsnames]{xcolor}
\usepackage[11pt]{moresize}
\usepackage{times,latexsym}
\usepackage{url}
\usepackage[T1]{fontenc}
\usepackage{longtable}
\usepackage{comment}
\usepackage{array}
\usepackage{float}
\usepackage{tacl2021v1}

\usepackage{xspace,mfirstuc,tabulary}
\usepackage{algorithm}
\usepackage{algpseudocode}
\usepackage{amsmath}
\usepackage{cleveref}
\usepackage{enumitem}
\usepackage{graphicx}
\usepackage{microtype}
\usepackage{multicol}
\usepackage{multirow}

\usepackage{amsfonts}
\usepackage{inconsolata}

\usepackage{color}
\usepackage{ colortbl}
\usepackage{booktabs}
\usepackage{algpseudocode}
\newif\iftaclinstructions
\taclinstructionsfalse 
\iftaclinstructions
\renewcommand{\confidential}{}
\renewcommand{\anonsubtext}{(No author info supplied here, for consistency with
TACL-submission anonymization requirements)}
\newcommand{\instr}
\fi

\taclpubformattrue
\AddToShipoutPicture{%
\AtPageLowishCenter{\thepage}
}

\iftaclpubformat 

\else

\fi 

\newcolumntype{L}[1]{>{\raggedright\let\newline\\\arraybackslash\hspace{0pt}}m{#1}}
\newcolumntype{C}[1]{>{\centering\let\newline\\\arraybackslash\hspace{0pt}}m{#1}}
\newcolumntype{R}[1]{>{\raggedleft\let\newline\\\arraybackslash\hspace{0pt}}m{#1}}

\newcommand{\snstatic}{\textsc{SN}$_\text{static}$}
\newcommand{\sndyna}{\textsc{SN}$_\text{dyna}$}

\newcommand{\nonep}{\textsc{NonEp}}
\newcommand{\meta}{\textsc{Meta}}

\definecolor{Gray}{gray}{0.88}
\definecolor{ceil}{rgb}{0.56, 0.6, 0.7}
\definecolor{blue}{rgb}{0.8, 0.85, 0.9}
\definecolor{grey}{rgb}{0.63, 0.65, 0.68}
\definecolor{purp}{rgb}{0.6, 0.7, 0.81}
\definecolor{orange}{rgb}{1.0, 0.6, 0.4}

\title{Data-Efficient Cross-Lingual Transfer with\\  Language-Specific Subnetworks}

\author{
  Rochelle Choenni
  \\
  University of Amsterdam
  \\
  \texttt{r.m.v.k.choenni@uva.nl}
  \And
  Dan Garrette 
  \\
  Google Research
  \\
  \texttt{dhgarrette@google.com}
   \And
  Ekaterina Shutova
  \\
  University of Amsterdam
  \\
  \texttt{e.shutova@uva.nl}
}

\date{}

\begin{document}
\maketitle

\begin{abstract}
Large multilingual language models typically share their parameters across all languages, which enables cross-lingual task transfer, but learning can also be hindered when training updates from different languages are in conflict. In this paper, we propose novel methods for using language-specific subnetworks, which control cross-lingual parameter sharing, to reduce conflicts and increase positive transfer during fine-tuning. We introduce dynamic subnetworks, which are jointly updated with the model, and we combine our methods with meta-learning, an established, but complementary, technique for improving cross-lingual transfer. Finally, we provide extensive analyses of how each of our methods affects the models.
\end{abstract}

\section{Introduction}

Large multilingual language models, such as mBERT \citep{devlin2018bert}, are pretrained on data covering many languages, but share their parameters across all languages. This modeling approach has several powerful advantages, such as allowing similar languages to exert positive influence on each other, and enabling cross-lingual task transfer (i.e., finetuning on some source language(s), then using the model on different target languages) \citep{pires2019multilingual}.
These advantages are particularly enticing in low-resource scenarios since without sufficient training data in the target language, the model's effectiveness hinges on its ability to derive benefit from other languages' data.
In practice, however, even state-of-the-art multilingual models tend to perform poorly on low-resource languages \citep{lauscher2020zero, ustun2020udapter}, due in part to \emph{negative interference} effects---parameter updates that help the model on one language, but harm its ability to handle another---which undercut the benefits of multilingual modeling \citep{arivazhagan2019massively, wang2020negative, ansell2021mad}.

In this paper, we propose novel methods for using language-specific subnetworks, which control cross-lingual parameter sharing, to reduce conflicts and increase positive transfer during fine-tuning, with the goal of improving the performance of multilingual language models on low-resource languages.
While recent works apply various subnetwork based approaches to their models statically \citep{lu2022language, yang2022learning, nooralahzadeh2022improving}, we propose a new method that allows the model to dynamically update the subnetworks during fine-tuning. This allows for sharing between language pairs to a different extent at the different learning stages of the models. We accomplish this by using pruning techniques \citep{frankle2018lottery} to select an optimal subset of parameters from the full model for further language-specific fine-tuning. Inspired by studies that show that attention-heads in BERT-based models have specialized functions \citep{voita2019analyzing, htut2019attention}, we focus on learning subnetworks at the attention-head level. We learn separate---but potentially overlapping---head masks for each language by fine-tuning the model on the language, and then pruning out the least important heads.

Given our focus on low-resource languages, we also combine our methods with meta-learning, a data-efficient technique to learn tasks from a few samples \citep{finn2017model}. Motivated by  \citet{wang2020negative}, who find that meta-learning can reduce negative interference in the multilingual setup, we test how much our subnetwork methods can further benefit performance in this learning framework, as well as compare the subnetwork based approach to a meta-learning baseline. 
Our results show that a combination of meta-learning and dynamic subnetworks is especially powerful.
To the best of our knowledge, we are the first to adapt subnetwork sharing to the meta-learning framework.

We extensively test the effectiveness of our methods on the task of dependency parsing. We use data from Universal Dependencies (UD) \citep{nivre2016universal} comprising 86 datasets covering 74 distinct languages, from 43 language families; 58 of the languages can be considered truly low-resource.  
Our experiments show, quantitatively, that our language-specific subnetworks, when used 
during fine-tuning, act as an effective sharing mechanism: permitting positive influence from similar languages, while shielding each language's parameters from negative interference that would otherwise have been introduced by more distant languages. 
Moreover, we show substantial improvements in cross-lingual transfer to new languages at test time. Importantly, we are able to achieve this while requiring considerably less time and data, i.e. training for less than a day compared to $\sim$20 days for current state-of-the-art \citep{kondratyuk201975}, while relying on data from just 8 treebanks.

Finally, we perform extensive analyses of our models to better understand how different choices affect generalisation properties. We analyse model behaviour with respect to several factors: typological relatedness of fine-tuning and test languages, data-scarcity during pretraining, robustness to domain transfer, and their ability to predict rare and unseen labels. We find interesting differences in model behaviour that can provide useful guidance on which method to choose based on the properties of the target language.

\section{Background and related work}

\subsection{Pruning and sparse networks}

\citet{frankle2018lottery} were the first to show that
neural network pruning \citep{han2015learning, li2016pruning} can be used to find a subnetwork that
matches the test accuracy of the full network. Later studies confirmed that such subnetworks also exist within BERT \citep{prasanna2020bert}, and that they can even be transferred across different NLP tasks \citep{chen2020lottery}. While these studies are typically motivated by 
a desire to find a smaller, 
faster version of the model \citep{jiao2020tinybert, lan2019albert, sanh2019distilbert}, we use pruning to find multiple simultaneous subnetworks (one for each fine-tuning language) within the overall multilingual model, which we use during both fine-tuning and inference to guide cross-lingual sharing.

\subsection{Selective parameter sharing}
\citet{naseem2012selective} used categorizations from linguistic typology to explicitly share subsets of parameters across separate languages' dependency parsing models.
Large multilingual models 
have, however, been shown to induce implicit typological properties automatically, and different design decisions (e.g., training strategy) can influence the language relationships they encode \citep{chi2020finding, choenni2022investigating}. 
Rather than attempting to force the model to follow an externally defined typology,
we instead take a data-driven approach, using pruning methods to automatically identify the subnetwork of parameters most relevant to each language, and letting subnetwork overlap naturally dictate parameter sharing.

A related line of research aims to control selective sharing by injecting language-specific parameters \citep{ustun2020udapter, wang2020negative, le2021lightweight, ansell2021mad}, which is often realized by inserting adapter modules into the network \citep{houlsby2019parameter}.
Our approach, in contrast, uses subnetwork masking of the existing model parameters to control language interaction.

Lastly, \citet{wang2020negative} separate language-specific and language-universal parameters within \textit{bilingual} models, and then meta-train the language-specific parameters only. However, given that we work in a multilingual as opposed to a bilingual setting, most parameters are shared by at least a few languages, and are thus somewhere between purely language-specific and fully universal.
Our approach, instead, allows for parameters to be shared among any specific subset of languages.

\paragraph{Analyzing and training shared subnetworks}
The idea of sharing through sparse subnetworks was first proposed for multi-task learning \citep{sun2020learning} and was recently studied in the multilingual setting: \citet{foroutan2022discovering} show that both language-neutral and language-specific subnetworks exist in multilingual models, and \citet{nooralahzadeh2022improving} show that training \textit{task-specific} subnetworks can help in cross-lingual transfer as well. In concurrent work, \citet{lu2022language} show that using language-specific subnetworks at the pretraining stage can mitigate negative interference 
for speech recognition.
We instead apply subnetworks during fine-tuning and few-shot fine-tuning at test time, allowing us to both make use of existing pretrained models and apply our models to truly low-resource languages. Moreover, we go beyond existing work by experimenting with \emph{structured} subnetworks, by allowing subnetworks to dynamically change during fine-tuning, 
and by extensively analyzing the effects and benefits of our methods.

\subsection{Meta-learning}

Meta-learning is motivated by the idea that a model can `learn to learn' many tasks from only a few samples. This has been adapted to the multilingual setting by optimising a model to be able to quickly adapt to new languages: by using meta-learning to fine-tune a multilingual model on a small set of (higher-resource) languages, the model can then be adapted to a new language using only a few examples \citep{nooralahzadeh2020zero}.
In this work, we use the Model-Agnostic Meta-Learning algorithm (MAML) \citep{finn2017model}, which has already proven useful for cross-lingual transfer of NLP tasks \citep{nooralahzadeh2020zero, wu2020enhanced, gu2020meta}, including being applied to dependency parsing by \citet{langedijk2022meta}, whose approach we follow for our own experiments.

MAML iteratively selects a batch of training tasks $\mathcal{T}$, also known as \textit{episodes}. For each task $t\in\mathcal{T}$, we sample a training dataset  $\mathcal{D}_{t} = (\mathcal{D}^\textit{trn}_{t} \cup \mathcal{D}^\textit{tst}_{t})$ that consists of a \textit{support set} used for adaptation, and a \textit{query set} used for evaluation.
MAML casts the meta-training step as a bilevel optimization
problem. Within each episode, the parameters $\theta$ of a model $f_{\theta}$ are fine-tuned on the support set of each task $t$ yielding $f_{\phi_{t}}$, i.e.\ the model adapts to a new task. The model $f_{\phi_{t}}$ is then evaluated on the query set of task $t$, for all of the tasks in the batch. This adaptation step is referred to as the \textit{inner loop} of MAML. In the \textit{outer loop},  the original model $f_{\theta}$ is then updated using the gradients of the query set of each $t\in\mathcal{T}$ with respect to the original model parameters $\theta$. 
 MAML strives to learn a good initialisation of $f_{\theta}$, which allows for quick adaptation to new tasks. This setup is mimicked at test time where we again select a support set from the test task for few-shot adaptation, prior to evaluating the model on the remainder of the task data.  

\subsection{Dependency parsing}
In dependency parsing, a model must predict, given an input sentence, a \emph{dependency tree}: a directed graphs of binary, asymmetrical arcs between words.
Each arc is labeled with a dependency relation type that holds between the two words, commonly referred to as the \textit{head} and its \textit{dependent}. 

The Universal Dependencies (UD) project has brought forth a dependency formalism that allows for consistent morphosyntactic annotation across typologically diverse languages \citep{nivre2016universal}. While UD parsing has received much attention in the NLP community, performance on low-resource languages remains far below that of high-resource languages \citep{zeman2018conll}. State-of-the-art multilingual parsers exploit pre-trained mBERT with a deep biaffine parser \citep{dozat2016deep} on top. The model is then fine-tuned on data (typically) from high-resource languages. This fine-tuning stage has been performed on English data only \citep{wu2020enhanced}, or multiple languages \citep{tran2019zero}.
UDify \citep{kondratyuk201975} takes this a step further and is fine-tuned on all available training sets together (covering 75 languages). Moreover, they use a multi-task training objective that combines parsing with predicting part-of-speech tags, morphological features, and lemmas.
UDapter \citep{ustun2020udapter} is trained on 13 languages using the same setup as UDify, but freezes mBERT's parameters and trains language-specific adapter modules. Moreover, they induce typological guidance by taking language embeddings predicted from typological features as input.

\section{Data} \label{ref:task-data} 

We use data from Universal Dependencies v2.9\footnote{https://universaldependencies.org/} and test on 86 datasets covering 74 unique and highly typologically diverse languages belonging to 21 language families from 43 subfamilies. We consider 58 of these languages to be truly low-resource as there are fewer than 31 training samples available. For the other 28 languages, 50\% have approximately 150--2K training samples and the other 50\% have 2K--15K samples available. In total, our test data contains 233 possible arc labels. We use 8 high-resource languages for fine-tuning, based on the selection used by \citet{langedijk2022meta} and \citet{tran2019zero}: English, Arabic, Czech, Estonian, Hindi, Italian, Norwegian, and Russian. 

\section{Methodology}\label{ref:methodology}
In \S\ref{ref:model}--\ref{ref:train} we describe the model that will be used throughout our experiments and the training strategy. In \S\ref{ref:subnetworks} we then explain how we define and select subnetworks, and how we apply them to our models. In \S\ref{ref:meta} we explain how our approach is adapted to the meta-learning setting, and in \S\ref{ref:fewshot}--\ref{ref:baselines} we describe our test setup and baselines. 

\subsection{Model}\label{ref:model}
Our implementation is derived from UDify \citep{kondratyuk201975}, but uses only the parsing task rather than its full multi-task setup. The model is built on mBERT \citep{devlin2018bert}, a bidirectional Transformer \citep{vaswani2017attention} with 12 layers, each with 12 attention heads, pretrained on the combined Wikipedia dumps of 104 languages, and using a shared WordPiece vocabulary for tokenization. 
We initialise the model from  mBERT, plus random initialization of the task-specific classifier. For each input token $j$, a weighted sum $r_j$ over all layers $i \in [1..12]$ is computed as follows:
\begin{equation}
    r_j = \eta \sum_{i} \mathbf{U}_{i,j} \cdot \text{softmax}(\lambda)_i
\end{equation}
where 
$\mathbf{U}_{i,j}$ is the output of layer $i$ at token position $j$,
$\lambda$ is a vector of trainable scalar mixing weights that distribute importance across the layers, and $\eta$ is a trainable scalar that scales the normalized averages.
For words that were tokenized into multiple word pieces, only the first word piece is used as input to the task-specific graph-based biaffine attention classifier \citep{dozat2016deep}. 

The classifier projects the word encodings $r_j$ through separate arc-head and arc-child feedforward layers with 768 hidden dimensions and  Exponential Linear Unit (ELU) non-linear activation. The resulting outputs $H_\text{arc-head}$ and $H_\text{arc-dep}$ are then combined using the biaffine attention function with weights $\mathbf{W}_\text{arc}$ and bias  $\mathbf{b}_\text{arc}$ to score all possible dependency arcs: 
\begin{equation}
S_\text{arc}=H_\text{arc-head} \mathbf{W}_\text{arc} H^{T}_\text{arc-dep}+\mathbf{b}_\text{arc}
\end{equation}
Similarly, we compute label scores $S_\text{tag}$ by using another biaffine attention function over two separate tag-head and tag-child feedforward layers with 256 hidden dimensions. The Chu-Liu/Edmonds algorithm \citep{chu1965shortest} is then used to select the optimal valid candidate tree.

\subsection{Training procedure}\label{ref:train}
Taking inspiration from \citet{nooralahzadeh2020zero} for cross-lingual transfer to low-resource languages, our training procedure is split into two stages: (1) fine-tune on the full English training set ($\sim$12.5K samples), without applying any subnetwork restrictions, for 60 epochs, to provide the full model with a general understanding of the task; and (2) fine-tune on the 7 other high-resource languages, to give the model a broad view over a typologically diverse set of languages in order to facilitate cross-lingual transfer to new languages.

For stage 2, in each iteration, we sample a batch from each language and average the losses of all languages to update the model. During this stage, we restrict each example to just the parameters in that language's subnetwork.
We perform 1000 iterations, with a batch of size 20 from each of the 7 languages, for a total of 140K samples.

We use a cosine-based learning rate scheduler with 10\% warm-up and the Adam optimizer \citep{kingma2014adam}, with separate learning rates for updating the encoder and the classifier (see Appendix~\ref{app}, Table~\ref{Table:hp} for 
details).

\subsection{Subnetwork masks} \label{ref:subnetworks}

We represent language-specific subnetworks as masks that are applied to the model in order to ensure that only a subset of the model's parameters are activated (or updated) during fine-tuning and inference.
We follow \citet{prasanna2020bert} in using \emph{structured} masks, treating entire attention heads as units which are always fully enabled or disabled. Thus, for language $\ell$, its subnetwork is implemented as a binary mask $\xi_\ell \in \{0, 1\}^{12 \times 12}$.

In our experiments, we present two ways of using the masks during fine-tuning: \textit{statically}, in which we find initial masks based on the pretrained model parameters and hold those masks fixed throughout fine-tuning and inference (\snstatic{}); and \textit{dynamically}, in which we update those masks over the course of fine-tuning (\sndyna{}).

\subsubsection{Finding initial subnetwork masks}\label{ref:mask-init}

We aim to find a mask for each of the 7 fine-tuning languages that prunes away as many heads as possible without harming performance for that language (i.e., by pruning away heads that are only used by other languages, or that are unrelated to the dependency parsing task).
For this, we apply the procedure introduced by \citet{michel2019sixteen}. 

For a language $\ell$, the procedure starts by fine-tuning the model on $\ell$'s training set. We then iterate by repeatedly removing the 10\% of heads with the lowest importance scores $\text{HI}^{(i,j)}_\ell$ ($i$=head, $j$=layer), which is estimated based on the expected sensitivity of the model to mask variable $\xi^{(i,j)}_\ell$:
\begin{equation}
     \text{HI}^{(i,j)}_\ell = \mathop{\mathbb{E}_{x_\ell \sim X_\ell}} \left | \frac{\delta \mathcal{L}(x_\ell)}{\delta \xi^{(i,j)}_\ell} \right |
\end{equation}
where $X_\ell$ is $\ell$'s data distribution,
$x_\ell$ is a sample from that distribution, and 
$\mathcal{L}(x_\ell)$ is the loss with respect to the sample. The procedure stops when performance on the $\ell$'s development set reaches 95\% of the original model performance.

Consistent with findings from \citet{prasanna2020bert}, we observed that the subnetworks found by the procedure are unstable across different random initializations. 
To ensure that the subnetwork we end up with is more robust to these variations, we repeat the pruning procedure with 4 random seeds, and take the union\footnote{Stricter criteria (e.g.\ the intersection of the 4 subnetworks) resulted in lower performance on the development set.} of their results as the true subnetwork (i.e., it includes even those heads that were only \emph{sometimes} found to be important).


\subsubsection{Dynamically adapting subnetworks}

\citet{blevins2022analyzing} showed that multilingual models acquire linguistic knowledge progressively---lower-level syntax is learned prior to higher-level syntax, and then semantics---but that the order in which the model learns to transfer information between specific languages varies.
As such, the optimal set of parameters to share may depend on what learning stage the model is in, or on other factors, e.g.\ the domains of the specific training datasets, the amounts of data available, the complexity of the language with respect to the task, etc.
Thus, we propose a dynamic approach to subnetwork sharing, in which each language's subnetwork mask is trained jointly with the model during fine-tuning.
This allows the subnetwork masks to be improved, and also allows for different patterns of sharing at different points during fine-tuning.

For dynamic adaptation, we initialise the identified static subnetworks as described in \S\ref{ref:mask-init} using small positive weights. We then allow the model to update the mask weights during fine-tuning. After each iteration, the learned weights are fed to a threshold function that sets the smallest 20$\%$ of weights to zero (i.e.\ 28 heads) to obtain a binary mask again. Given that the derivative of a threshold function is zero, we use a straight-through estimator \citep{bengio2013estimating} in the backward pass, meaning that we ignore the derivative of the threshold function and pass the incoming gradient on as if the threshold function was an identity function. 
\algnewcommand{\parState}[2]{\State%
  \parbox[t]{\dimexpr\linewidth-\algorithmicindent*#1}{\strut  #2\strut}}

\algnewcommand{\Initialize}[1]{%
  \State \textbf{Initialize:} \hspace*{\algorithmicindent}\parbox[t]{.8\linewidth}{\raggedright #1}
}

\begin{algorithm}[t!]
\caption{Meta-training procedure}\label{alg:cap}
\begin{algorithmic}
\Require{Language datasets $\mathcal{T}$; step sizes $\alpha$ and $\beta$; number of updates $k$; number of episodes \texttt{EPS}; support/query set size $N$; and subnetworks $\{\xi_\ell \mid \ell \in \mathcal{T}\}$.}
Train on $\ell \notin \mathcal{T}$ to yield initial parameters $\theta$.
\For{\texttt{EPS}}:
    \For{$\ell \in \mathcal{T}$} :\hspace{2.4cm} (\textit{\textbf{inner loop}})
        \State{Yield learner: $\phi_\ell \gets \theta$.copy()}
        \State{Mask $\phi_\ell$ using $\xi_\ell$}
        \State{Take $N$ samples to form $\mathcal{D}^\textit{trn}_\ell=\{x\}^N_{n=1}$} \State{$ \in \mathcal{T}_\ell$ and $\mathcal{D}^\textit{tst}_\ell = \{x\}^N_{n=1} \in  \mathcal{T}_\ell$}
        \State{Update on the \textit{support set}:}
        \For{$k$ steps}:
            \State{$\phi_{\ell} \gets \theta -\alpha \nabla_{\theta} \mathcal{L}(\phi_\ell, \mathcal{D}^\textit{trn}_\ell )$}
        \EndFor
        \State{Evaluate on the \textit{query set}: $\mathcal{L}(\phi_\ell, \mathcal{D}^\textit{tst}_\ell  )$}
    \EndFor
    \parState{1}{Meta-update: \hspace{2.4cm} (\textit{\textbf{outer loop}})\\
    $\theta \gets \theta - \beta \sum_{\ell\in \mathcal{T}} \nabla_{\theta} \mathcal{L}(\phi_\ell, \mathcal{D}^\textit{tst}_\ell)$}
\EndFor
\end{algorithmic}
\end{algorithm}

\subsection{Meta-learning with subnetworks}\label{ref:meta}
Meta-learning for multilingual models has been shown to enable both quick adaptation to unseen languages \citep{langedijk2022meta} and mitigation of negative interference \citep{wang2020negative}, but it does so using techniques that are different from---though compatible with---our subnetwork-sharing approach. Therefore, we experiment with the combination of these methods, and test the extent to which their benefits are complementary (as opposed to redundant) in practice.

To integrate our subnetworks within a meta-learning setup, we just have to apply them in the inner loop of MAML, i.e.\ given a model $f$ parameterised by $\theta$, we train $\theta$ by optimizing for the performance of the learner model of a language $\ell$ masked with the corresponding subnetwork $f_{\phi_{\ell}} \cdot \xi_{l}$. See Algorithm 1 for the details of the procedure.\footnote{Note that for the meta-update, we use a first-order approximation, replacing $\nabla_\theta\mathcal{L}(\phi_\ell, \mathcal{D}^\textit{tst}_\ell) $ by $\nabla_{\phi}\mathcal{L}(\phi_\ell, \mathcal{D}^\textit{tst}_\ell) $. See \citet{finn2017model} for more details on first-order MAML.}

For all meta-learning experiments, we train for 500 episodes with support and query sets of size 20, i.e.\ 10K samples per language are used for meta-training and validation each. We use 20 inner loop updates ($k$) and we follow \citet{finn2017model} in using SGD for updating the learner. All other training details are kept consistent with the non-episodic (\nonep{}) models (as described in \S\ref{ref:train}).

\subsection{Few-shot fine-tuning at test time}\label{ref:fewshot}

Since the primary goal of this work is to improve performance in low-resource scenarios, we evaluate our models using a setup that is appropriate when there is almost no annotated data in the target language: few-shot fine-tuning. For a given test language, the model is fine-tuned on just 20 examples in that language, using 20 gradient updates.  
The examples are drawn from the development set, if there is one; 
otherwise they are drawn from (and removed from) the test set.
We use the same hyperparameter values as during training.
We report Labeled Attachment Scores (LAS) averaged across 5 random seeds, as computed by the official CoNLL 2018 Shared Task evaluation script.\footnote{https://universaldependencies.org/conll18/evaluation.html}

Since we do not have subnetworks for the test languages---only for the 7 high-resource languages used in stage 2 of fine-tuning (\S\ref{ref:train})---we instead use the subnetwork of the typologically most similar training language.
We determine typological similarity by computing the cosine similarity between the language vectors from the URIEL database (\texttt{syntax\_knn}) \citep{littell2017uriel}.

\begin{table}[!t]
    \begin{tabular}{l@{\hspace{1.5\tabcolsep}}l@{\hspace{1.0\tabcolsep}}r@{\hspace{1.0\tabcolsep}}r@{\hspace{1.0\tabcolsep}}r|@{\hspace{0.75\tabcolsep}}r@{\hspace{0.25\tabcolsep}}}
    \toprule
      & & \textsc{Full} & \snstatic{} & \sndyna{} & Total\\ 
     \midrule
     \multirow{2}*{\nonep{}}
      & LAS    & 37.6 & \textbf{40.4} & 39.0 \\
      & Best\% &  1\% & 23\% & 8\% & 33\%\\
      \multirow{2}*{\textsc{Meta}}
      & LAS    & 39.7 & 39.3 & \textbf{39.9}\\
      & Best\% & 14\% & 26\% & \textbf{28\%} & \textbf{67\%} \\
    \midrule
\end{tabular}
\centering
\vspace{-0.2cm}
\caption{Results 
on UD Parsing, for both non-episodic (\nonep{}) and meta-learning (\textsc{Meta}) setups. For each of the 6 models, we report Labeled Attachment Score (LAS) averaged across all 86 test languages, as well as the percentage of languages for which that model performed best (e.g., \meta{}-\sndyna{} yielded the highest LAS on 28\% of test languages).}
\label{Table:winning}
\end{table}

\subsection{Baselines}\label{ref:baselines}

To measure the effectiveness of our subnetwork-based methods, we train and evaluate baselines in which no subnetwork masking is applied (but for which all other details of the training and testing setups are kept unchanged). We refer to this as \emph{full model training} (\textsc{Full}) to contrast our training approaches that use static or dynamic subnetworks (\snstatic{} and \sndyna{}), and we report these baselines for both the non-episodic (\nonep{}) and meta-learning (\meta{}) frameworks.

\section{Results}\label{sec:avgres}

\begin{figure}[t!]
    \centering
    \includegraphics[width=\linewidth, height=3.5cm]{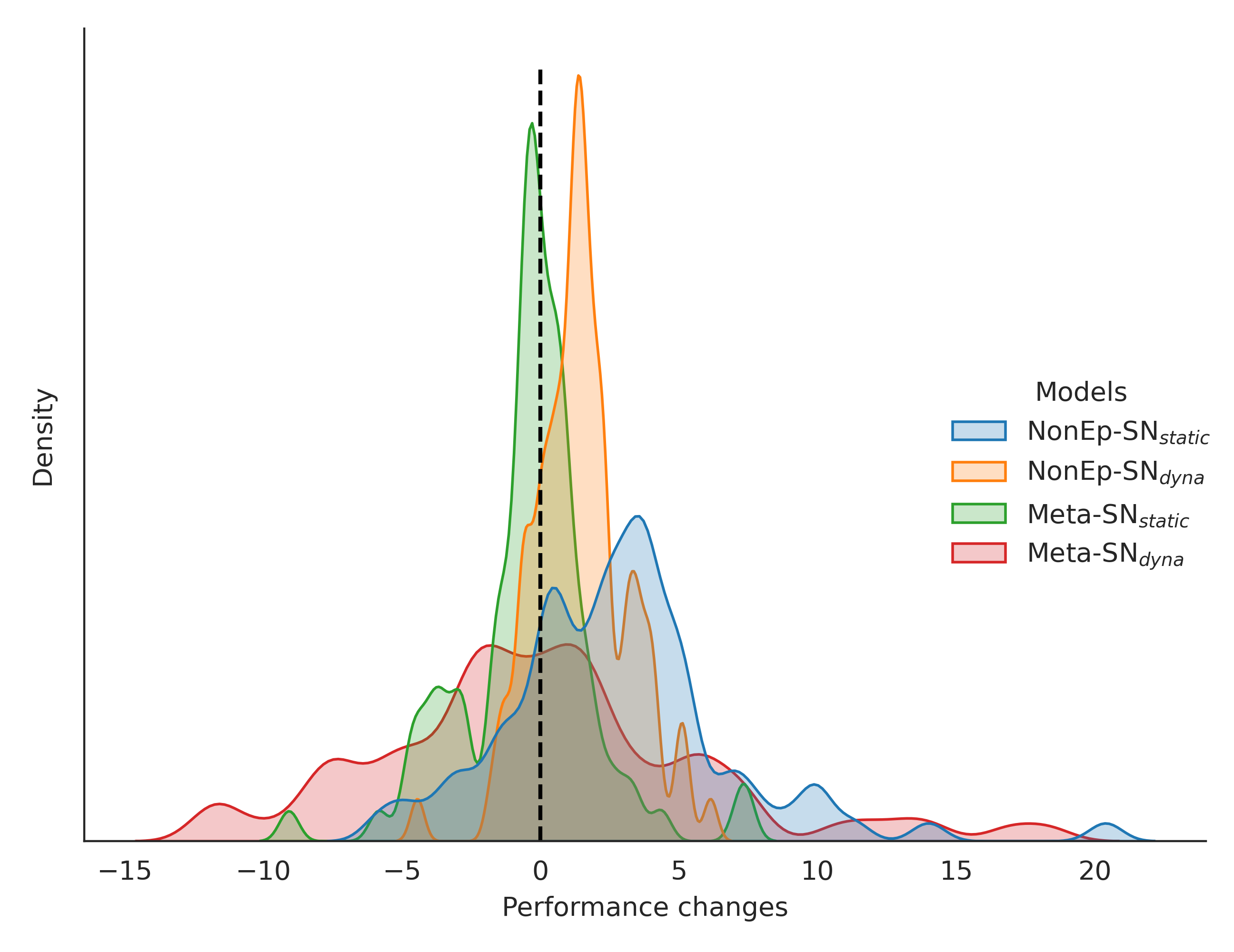}
    \vspace{-0.4cm}
    \caption{Kernel density estimation (KDE) plot over the relative performance changes of each model for all test languages when comparing to its corresponding full model training baseline. }
    \label{fig:kde}
\end{figure}

Overall, the results show that our subnetwork-based methods yield improvements over baseline models trained without any subnetwork masking.
In Table \ref{Table:winning}, we see that, based on average LAS scores across all test languages, static subnetworks (\snstatic) perform best in the non-episodic training setup, resulting in +2.8\% average improvement over the \textsc{Full} baseline, and yielding the highest average LAS of all the models.
Dynamic subnetworks (\sndyna{}), on the other hand, exhibit superior performance in the meta-learning setting, resulting in the model that performed best for the largest number of languages.
In Table \ref{Table:hm_res}, we report the full set of results on all 86 test languages. 

To gain more insight into the effects of our methods across the test languages, we plot the distribution over performance changes compared to the baseline per method and learning framework in Figure \ref{fig:kde}. We find that static and dynamic subnetworks exhibit opposite trends. \nonep{}-\snstatic{} achieves large gains (up to +25\%), but can also cause more deterioration on other languages (up to $-$6\%). In contrast, the performance change distribution for \nonep{}-\sndyna{} is centered around more modest improvements, but is also the safest option given that it deteriorates performance for the fewest languages. The same trade-off can be observed in the meta-learning framework, except that now \meta{}-\snstatic{} results in modest changes compared to \meta{}-\sndyna{}.


Lastly, we do not find strong trends for transfer languages;\ different magnitudes of performance changes are scattered across all transfer languages.

\begin{table*}[htb]
\small
     \begin{tabular}
{p{4cm}|p{1cm}p{1cm}|p{1cm}p{1cm}p{1cm}|p{1cm}p{1cm}p{1cm}}
     \cline{4-9}
        \multicolumn{2}{c}{} &
     &  \multicolumn{3}{c|}{Non-Episodic (\nonep{})}  & \multicolumn{3}{c|}{Meta-Learning (\textsc{\meta{}})} \\
     \hline
     \cellcolor{grey} \textbf{From Arabic} ($\bar{ \theta}$ = 0.70)  & \cellcolor{purp} UDF & 
     \cellcolor{purp} UDA & \cellcolor{ceil} \textsc{Full} & \cellcolor{ceil} \snstatic{} & \cellcolor{ceil} \sndyna{} & \cellcolor{blue} \textsc{Full} & \cellcolor{blue} \snstatic{} & \cellcolor{blue} \sndyna{} \\
    \hline
    {Guajajara TuDeT}
        & - & - & 28.61 & 27.07 &  \colorbox{Goldenrod}{\textbf{33.62} } & {25.80} & {26.11} & \colorbox{orange}{\textbf{30.90}}\\
        \hline
     {Kiche IU} 
         & - & - & 40.13 & \textbf{41.24} & 40.04 & 30.10 & 21.03 & \colorbox{BlueGreen}{\textbf{40.78}}\\
        \hline    
         {Indonesian GSD}
     & 80.10& - & 63.64 & 63.96 & \textbf{64.05} & 62.39 & 62.42 & \textbf{64.73}\\
        \hline
         {Indonesian PUD}
         & 56.90 & - & {70.08} & \colorbox{orange}{\underline{\textbf{74.88}}} & {71.54} &  \textbf{74.71} & 73.32 &71.64 \\
        \hline
         {Javanese CSUI} 
         & - & - & {57.80} & \colorbox{orange}{\textbf{61.92}} &{60.07} & 62.40 & \textbf{62.94} & 60.84 \\
        \hline
     {Maltese MUDT} 
         & 75.56 & - & \textbf{29.37} & 24.67 & 28.47 & {16.92} & {13.43} & \colorbox{BlueGreen}{\textbf{30.05}} \\    
        \hline
           {Mbya Guarani Thomas}
         & - & 8.40 & 16.76 & 16.30 & \underline{\textbf{17.61}} & {10.79} & {11.55} & \colorbox{Goldenrod}{\textbf{16.55}}\\
        \hline  
        {South Levantine Arabic} 
         & - & - & 39.42 & \textbf{41.93} & 41.2 & 42.05 & 42.32 & \textbf{42.37} \\
        \hline
        {Thai PUD} 
        &  - & 26.06  & {33.82} & \colorbox{Goldenrod}{\underline{\textbf{39.12}}} & \colorbox{orange}{36.94}  &  37.66 & \textbf{38.32} & 37.95 \\
        \hline
          {Tagalog
        TRG} 
         & 40.07 & 69.52 & 70.46 & 65.26& \textbf{71.82}  & \underline{\textbf{73.17}} & 71.58& 72.58\\
        \hline
        {Tagalog Ugnayan} 
         & - & - & 48.39 & 47.38 & \textbf{49.76} & 50.93 & 46.69 & \textbf{53.22}\\
        \hline
         {Vietnamese
        VTB} 
         & 66.0 & - & {40.79} & \colorbox{orange}{\textbf{44.62}} & {43.34}  &  \textbf{45.24} & 43.75 & 43.67\\
        \hline
        {Wolof
        WTB}
          & - & - & 20.72 & 18.98 & \textbf{22.55} & {17.26} &{15.56} & \colorbox{Goldenrod}{\textbf{24.63}}\\
         \hline
         \rowcolor{Gray}\textbf{Average (13)} & - & - & 43.07 & 43.64 & \textbf{44.69} & 42.26& 40.66 & \textbf{45.38}  \\
        \hline
       \rowcolor{grey} \textbf{From Czech}  ($\bar{\theta}$ = 0.83) & 
     \cellcolor{purp} UDF & 
     \cellcolor{purp} UDA & \cellcolor{ceil} \textsc{Full} & \cellcolor{ceil} \snstatic{} & \cellcolor{ceil} \sndyna{} & \cellcolor{blue} Meta & \cellcolor{blue} \snstatic{} & \cellcolor{blue} \sndyna{} \\
        \hline
       {Armenian
        ArmTDP}
         & 78.61 & - & {48.22}& \colorbox{BlueGreen}{\textbf{58.35}} & \colorbox{orange}{52.07} & {57.64}  &  \colorbox{orange}{\textbf{61.09}} & {50.66} \\
        \hline 
        {Armenian
        BSUT} 
         & - & - & {57.26} & \colorbox{Goldenrod}{\textbf{64.75}} & {59.49} & {62.95} & \colorbox{orange}{\textbf{67.30}}  & {60.24}\\
        \hline    
         {Kurmanji
        MG} 
         & 20.40  & 12.10 & {13.28} & \colorbox{orange}{\textbf{16.40}} & {14.78}  & 15.57 & 12.86 & \textbf{17.28}\\
        \hline
       {Lithuanian ALKSNIS} 
         & - & - & {50.09}& \colorbox{YellowGreen}{\textbf{59.98}} & \colorbox{YellowGreen}{57.28} & 60.81& \textbf{61.20} & 53.15\\
        \hline
        {Lithuanian
        HSE} 
          & 69.34 & - & {53.02} & \colorbox{Goldenrod}{\textbf{59.74}} &\colorbox{orange}{57.28}  &  61.26 &  \textbf{61.38} & 55.57\\
        \hline
         {Western Armenian} 
         & - & - & {43.01}  & \colorbox{BlueGreen}{\textbf{57.0} }&  \colorbox{Goldenrod}{49.14} &  56.93 & \textbf{58.34} & 48.62\\
        \hline
         \rowcolor{Gray} \textbf{Average (6)}  & & & 44.15&\textbf{52.7} & 48.34 & 52.53& \textbf{53.70} & 47.59  \\
        \hline
        \rowcolor{grey} \textbf{From Estonian}  ($\bar{\theta}$ = 0.84)  & 
        \cellcolor{purp} UDF & 
        \cellcolor{purp} UDA & \cellcolor{ceil} \textsc{Full} & \cellcolor{ceil} \snstatic{} & \cellcolor{ceil} \sndyna{} & \cellcolor{blue} Meta & \cellcolor{blue} \snstatic{} & \cellcolor{blue} \sndyna{} \\
        \hline
       {Apurina UFPA}
         & - & - & 37.70  & \textbf{39.66} & 37.68 & {28.18}   & {24.11}& \colorbox{YellowGreen}{\textbf{35.75}} \\
        \hline
        {Erzya
        JR}
         & 16.38 & 19.20 & 16.06 & 
        \textbf{17.35}& 16.39 & 17.77 & \textbf{18.66} &  15.64 \\
        \hline
        {Hungarian Szeged}
        & 84.88 & - & {53.38} & \colorbox{YellowGreen}{\textbf{62.24}} & {54.51} & {61.67} &  \colorbox{YellowGreen}{\textbf{68.69}} & {50.20}\\
        \hline
         {Karelian KKPP} 
         & - &48.40 & {36.67} & \colorbox{YellowGreen}{\textbf{43.69}} & {38.41} & 40.58 & 40.19 & \textbf{40.93}\\
        \hline
        {Komi Permyak UH} 
         & - & 23.10 & 24.47 & \textbf{26.19} & 25.86 & 24.96 & \underline{\textbf{26.52}} & 25.65\\
        \hline
        {Komi Zyrian IKDP} 
         & 22.12& - & 22.58 & \underline{\textbf{25.55}}& 23.62 & \textbf{24.97} & 22.23 & 24.63\\
        \hline
        {Komi Zyrian
        Lattice} 
         & 12.99 & - & 14.17 & \underline{\textbf{16.43}} & 14.23  & 14.72 & \textbf{15.30} & 13.62 \\
        \hline
        {Livvi KKPP} 
         & - & 43.30 & {34.22} & \colorbox{orange}{\textbf{38.0}}  & {32.45} & 36.52 & \textbf{37.08}  & 33.45\\
        \hline
         {Moksha JR} 
         & - &  26.60 & {15.20} & \colorbox{Goldenrod}{\textbf{20.18}} & {16.30} &  18.79 & \textbf{20.57} & 16.65\\
        \hline
         {North Sami Giella} 
         &67.13 &-& 14.05 & 14.69 & \textbf{14.75} & 11.74 & 11.90 & \colorbox{orange}{\textbf{16.51}}\\
        \hline
         {Skolt Sami-Giellagas} 
         & - & - & 26.49 & 26.20 & \textbf{27.83} & 21.84& 18.10 & \colorbox{Goldenrod}{\textbf{27.66}} \\
        \hline
        {Tatar NMCTT} 
         & - & - & 52.63 & \colorbox{orange}{\textbf{56.56}} & \colorbox{orange}{55.67} & 55.79 &  \colorbox{orange}{\textbf{58.90}} & 54.61\\
        \hline
        {Tupinamba TuDeT} 
         & - & - & 21.24 & 21.65 & \textbf{22.74} & 16.68 & 15.30 &  \colorbox{orange}{\textbf{20.12}}\\
        \hline
{Turkish PUD} 
         & 46.07 & -  & 47.0 &  \colorbox{orange}{\textbf{50.91}} & 49.34  & 50.58 & 50.01 & \underline{\textbf{52.63}}\\
        \hline
         {Turkish IMST} 
     & 67.44 & -  & {34.90} & \colorbox{Goldenrod}{\textbf{40.60}} & {35.99} &  40.87 & \textbf{41.81} & 36.32\\
        \hline
        \rowcolor{Gray} \textbf{Average (15)}  & -& -& 30.05 & \textbf{33.33} & 31.05 &31.04& \textbf{31.30} & 30.96 \\
     \hline
       \rowcolor{grey} \textbf{From Hindi}  ($\bar{\theta}$ = 0.74) &  
        \cellcolor{purp} UDF & 
        \cellcolor{purp} UDA & \cellcolor{ceil} \textsc{Full} & \cellcolor{ceil} \snstatic{} & \cellcolor{ceil} \sndyna{} & \cellcolor{blue} Meta & \cellcolor{blue} \snstatic{} & \cellcolor{blue} \sndyna{} \\
       \hline
        {Akuntsu TuDeT} 
        & - & - & \textbf{24.71}  & 21.86 & 23.97 &  21.76  &  21.29 &  \colorbox{orange}{\textbf{25.55}}\\
        \hline
       {Amharic ATT} 
         & 3.49 & 5.90 & 11.44  & \textbf{12.95} & 11.18 & 10.90 & \underline{\textbf{13.54}}  & 12.41 \\
        \hline
       {Bambara
        CRB} 
        & 8.60 & 8.10 & 21.94 & \textbf{22.54} & 21.47 & {17.79} & {18.09} &  \colorbox{Goldenrod}{\textbf{\underline{23.76}}}\\
        \hline
         {Basque
        BDT} 
        & 80.97 & - & 45.81 & 47.59 & \textbf{48.60}  & \textbf{52.81} & 52.52  & 45.71 \\
        \hline
             {Beja NSC}
         & - & - &   18.07 & 14.79 & \textbf{19.95} & {14.04} & {8.21} &  \colorbox{Goldenrod}{\textbf{19.87}}\\
        \hline
              {Bengali BRU}
         & - & - & {43.49}& \colorbox{Goldenrod}{\textbf{48.67}}  & {42.87} & \textbf{58.91} & 58.52  & 47.33\\
        \hline
         {Bhojpuri BHTB} 
         & 35.90 & 37.30 & 36.0 & \underline{\textbf{38.76}} & 38.24 & 36.72  &  \textbf{37.70}  & 35.71 \\
        \hline
        {Buryat
        BDT}  
       & 26.28 & 28.90 & 15.95 & 16.50 & \textbf{17.31} & 24.26  & 25.02  &   \colorbox{orange}{\textbf{27.79}} \\
        \hline
           {Kaapor TuDeT} 
        & - & - & 30.54 & \textbf{33.29} & 29.87 & 30.77 & 30.18 &  \textbf{32.75}\\
        \hline
        {Kangri KDTB} 
        & - & - &30.79 & \colorbox{orange}{\textbf{34.32}} & \colorbox{orange}{34.20} & \textbf{36.16} & 35.90 & 35.77\\
        \hline
           {Karo TuDeT} 
        & - &  -   &18.47 & 19.01 & \textbf{19.32} & 17.47 & 17.38 & \textbf{18.76} \\
        \hline
          {Kazakh
        KTB} 
        & 63.66 & 60.70 & {45.35} & \colorbox{Goldenrod}{\textbf{50.50}} & {47.07} &  53.92 & \textbf{54.70}  & 48.56\\
        \hline
     {Makurap TuDeT} 
         & - & - & 25.26 & \textbf{25.35} & 23.98 & {20.63} & {20.07}& \colorbox{YellowGreen}{\textbf{28.47}}\\
                \hline

\end{tabular}
\end{table*}
\begin{table*}[htb]
 \small
     \begin{tabular}{p{4cm}|p{1cm}p{1cm}|p{1cm}p{1cm}p{1cm}|p{1cm}p{1cm}p{1cm}}
         {Marathi UFAL} 
        & 67.72 & 44.40 & 37.96 & \colorbox{orange}{\textbf{41.46}}& 37.72 & \textbf{51.21}& 50.78 & 39.17\\
        \hline
        {Munduruku TuDeT} 
         & - &  -& \textbf{35.73} & 32.74 & 34.25 & {29.43} & {28.87}& \colorbox{YellowGreen}{\textbf{36.51}}\\
        \hline
        {Sanskrit
        UFAL} 
         & 18.56 & 22.20 & 18.63 & 19.70 & \textbf{19.74} & 21.58 & \textbf{22.18} & 19.0 \\
        \hline
      {Sanskrit Vedic} 
        & - & - & 12.96 & \textbf{13.09}& 12.43 & 12.51 & 12.40 & \textbf{12.71}\\
        \hline
        {Tamil MWTT} 
        & - & - &  63.11 & \textbf{65.34} & 61.86 & \textbf{72.39} & 72.04 & 64.76\\
        \hline
      {Tamil TTB} 
        & 71.29  &  45.76 & 46.66 & \colorbox{orange}{\textbf{51.48}} & 48.10 & 52.0 & \textbf{53.89} & 46.76\\
        \hline
         {Uyghur
        UDT} 
        & 48.80 & - & 20.78& 21.07& \textbf{21.41}  & \textbf{21.04} & 19.93 & 20.91\\
        \hline
         {Warlpiri
        UFAL} 
         &7.96 &12.10 & 55.91 & \colorbox{orange}{59.30}& \colorbox{orange}{\textbf{59.40}} & { 42.78} & { 42.67} & \colorbox{Goldenrod}{ \underline{\textbf{59.69}}}\\
        \hline
          {Xibe XDT} 
           & - & - & 12.61 & \textbf{13.06} & 12.88 & 12.70 & 11.07 & \textbf{13.75}   \\
        \hline
        {Yakut YKTDT} 
         & - & - & 30.85 & \colorbox{orange}{\textbf{35.06}} & \colorbox{orange}{34.73} & 32.4 &  32.54 & \textbf{33.99} \\
        \hline  
          {Yupik SLI} 
        & - & - & 12.73 & 11.31 & \textbf{13.28}  & 8.84 & {9.28} & \colorbox{Orchid}{\textbf{33.92}} \\
        \hline
\rowcolor{Gray} \textbf{Average (24)} & -&- & 29.82 & \textbf{31.42} & 30.58 & \textbf{32.73}& 32.55 & 32.07   \\
        \hline
       \rowcolor{grey} \textbf{From Italian} ($\bar{ \theta}$ = 0.85) & 
     \cellcolor{purp} UDF & 
     \cellcolor{purp} UDA & \cellcolor{ceil} \textsc{Full} & \cellcolor{ceil} \snstatic{} & \cellcolor{ceil} \sndyna{} & \cellcolor{blue} Meta & \cellcolor{blue} \snstatic{} & \cellcolor{blue} \sndyna{} \\
       \hline
        Akkadian PISANDUB 
    &4.54 & 8.20 & 17.33  & 11.42 & \textbf{18.44} & 7.44 & 9.30 &  \colorbox{BlueGreen}{\underline{\textbf{19.28}}} \\
    \hline
     Akkadian RIAO
     & - & - & 21.87  & 17.99 & \textbf{23.17}  & 12.89  & 9.33& \colorbox{BlueGreen}{\textbf{27.01}}\\
    \hline
       {Assyrin AS} 
         & 9.10& 14.30 & \underline{\textbf{20.53}} & 19.41 & 16.09 & 14.91 & 10.13 & \textbf{16.12} \\
        \hline
        {Breton
        KEB}  
         & 39.84 & 58.50 & 50.33  & \colorbox{BlueGreen}{\textbf{61.63}} & \colorbox{orange}{53.89} & 63.05 &  \underline{\textbf{64.31}} & 56.67
        \\
        \hline
          {Galician TreeGal} 
         & 76.77 & - & 75.81 & \textbf{77.63} & 76.05 &  \textbf{\underline{78.41}} &77.95 & 76.09\\
        \hline
       
        {Greek GDT}
         & 92.15 & - &77.90 & \colorbox{orange}{\textbf{81.46}} & 80.14 & \textbf{81.13} & 80.79&  79.78\\
        \hline
         {Irish IDT}
         & 69.28 & - & 47.14 & \colorbox{orange}{\textbf{50.80}} & 49.04  & 52.03 & \textbf{53.10} & 49.42\\
        \hline
          {Ligurian GLT} 
         & - & - & 29.43 & \colorbox{Orchid}{\textbf{49.81}} &  \colorbox{orange}{34.05} & 44.26 & \textbf{46.65} & 34.30 \\
        \hline
        {Manx Cadhan}
         & - & - & 46.13 &  44.97& \textbf{46.64} & 40.52 & 36.70 & \colorbox{Goldenrod}{ \textbf{47.31}} \\
                \hline
         {Naija NSC} 
         &32.16 & 36.70& 32.0 & \colorbox{orange}{\textbf{35.59}} & 32.09 & 34.52 & 33.46&  \textbf{\underline{37.84}}\\
        \hline
        {Scottish Gaelic ARCOSG}
        & - & -& 15.41 & \colorbox{YellowGreen}{\textbf{23.28}} & \colorbox{orange}{18.57}  &  24.49 & \textbf{25.88} & 19.86\\
        \hline
         {Welsh CCG} 
         & - & 54.40 & 47.37 &  \colorbox{orange}{51.72} & \colorbox{Goldenrod}{\textbf{52.60}}  & \underline{\textbf{54.97}} & 53.18 & 51.10  \\
        \hline
        
        \rowcolor{Gray} \textbf{Average (12)} & & & 40.1 & \textbf{43.81} & 41.77 & 42.38& 41.73 & \textbf{42.90}   \\
        \hline
        \rowcolor{grey} \textbf{From Norwegian} ($\bar{ \theta}$ = 0.91) &  
     \cellcolor{purp} UDF & 
     \cellcolor{purp} UDA & \cellcolor{ceil} \textsc{Full} & \cellcolor{ceil} \snstatic{} & \cellcolor{ceil} \sndyna{} & \cellcolor{blue} Meta & \cellcolor{blue} \snstatic{} & \cellcolor{blue} \sndyna{} \\
        \hline
       {Afrikaans AfriBooms} 
         & - & - & 65.88 &63.57 & \textbf{65.94} &68.28 & 63.64 & \textbf{69.75}  \\
        \hline
         {Albanian TSA}  
         & - & - &  70.95 & \colorbox{Goldenrod}{\textbf{76.06}} & 73.34 &  \textbf{79.76} & 76.65  & 74.13\\
        \hline   
         {Faroese FarPaHC}
          & - & - &47.51 & 50.03 & \colorbox{orange}{\textbf{51.0}} & 49.24 & 44.70 & \colorbox{orange}{\textbf{54.13}}\\
    \hline
     {Faroese
        OFT} 
         & 59.26  & 69.2 & 60.95 & \colorbox{BlueGreen}{\underline{\textbf{70.36}}} & 63.76  & \textbf{70.12} & 69.41 & 65.87\ \\
        \hline
         {Gothic PROIEL} 
        & 79.37 & - & 19.23& \textbf{19.67} & 18.68 & 16.65 & 15.85 & \colorbox{orange}{\textbf{20.24}}\\
        \hline
        {Icelandic Modern}
         & - & - & 45.98 & \colorbox{orange}{\textbf{49.98}} & 47.44 & \textbf{53.45} & 50.43 &  51.27\\
        \hline
          {Low Saxon LSDC} 
         & - & - & 47.75 & \colorbox{orange}{\textbf{51.42}} & 49.88 & \textbf{50.26}  & 47.50& 50.08\\
        \hline
          {Swiss German UZH} 
        & - & 45.50 & 45.98 & \colorbox{YellowGreen}{\underline{\textbf{52.66}}} & 47.57 & \textbf{51.93} & 51.73 & 51.16\\
        \hline   
        \rowcolor{Gray} \textbf{Average (8)} & -& -& 50.48 &\textbf{54.22} & 52.20  & \textbf{54.96}& 52.49 & 54.58\\
        \hline
        \rowcolor{grey} \textbf{From Russian} ($\bar{ \theta}$ = 0.76)&  
     \cellcolor{purp} UDF & 
     \cellcolor{purp} UDA & \cellcolor{ceil} \textsc{Full} & \cellcolor{ceil} \snstatic{} & \cellcolor{ceil} \sndyna{} & \cellcolor{blue} Meta & \cellcolor{blue} \snstatic{} & \cellcolor{blue} \sndyna{} \\
       \hline
          {Ancient Greek PROIEL}
        & 82.11 & 72.66 &  23.68 & \colorbox{orange}{\textbf{28.38}}&  \colorbox{orange}{27.72}  &  23.81 & \colorbox{BlueGreen}{ \textbf{31.24}} & 26.32\\
        \hline    
   {Cantonese HK}
         & 32.01  & 32.80 & 28.66 & \textbf{31.17} & 30.58  & \underline{\textbf{33.02}}& 32.87  & 31.50\\
        \hline
         {Chinese CFL}  
         & 42.48  & - & 45.48 & \colorbox{orange}{\textbf{49.28}}& 48.26  & 49.88 &\underline{\textbf{50.67}}  & 47.50 \\
        \hline
         {Chinese HK}  
         & 49.32  & - & 47.20 & \textbf{49.94} & 48.26 & 52.31 &  \underline{\textbf{52.90}} & 48.16  \\
        \hline
         {Chinese PUD}  
         & 56.51  & - & 44.74  & \textbf{46.98} & 46.47 & 45.9 & 44.92  & \textbf{ 46.71} \\
        \hline
          {Serbian
        SET} 
          & 91.95 & - & 78.98 & \textbf{81.57} & 79.66 & \textbf{80.98} & 80.96 & 79.8 \\
        \hline
        {Upper Sorbian
        UFAL} 
         & 62.82 & 54.20 & 49.81 & \colorbox{Goldenrod}{\textbf{54.88}} & \colorbox{orange}{53.84} & \textbf{54.01} & 53.78 & 51.29\\
        \hline
          {Yoruba YTB} 
        & 19.09 & 42.70 & 38.11 & 38.17 & \textbf{38.75}  & 38.79 & 38.17 & \textbf{39.28}\\
        \hline
         \rowcolor{Gray}\textbf{Average (8)} & - & - & 44.58& \textbf{47.55} & 46.62 & 47.34& \textbf{48.19} & 46.32 \\
        \hline
         \rowcolor{grey}\textbf{Total Avg. (86)} & -& -& 37.61 & \textbf{40.38} & 39.04  & 39.67 & 39.25 & \textbf{39.92}\\
         \bottomrule
\end{tabular}
\caption{Average LAS scores across 5 random seeds for all test languages (we omit reporting standard deviations as they were overall very small (6e-05--0.09)). We highlight subnetwork-based models that substantially improve over their full-model baselines, and color-code based on the amount of improvement: 
\colorbox{orange}{~} +3--5\%, \colorbox{Goldenrod}{~} +5--7 \%, \colorbox{YellowGreen}{~} +7--10 \%, \colorbox{BlueGreen}{~} +10--15\%, \colorbox{Orchid}{~} +20--25\%. Results are grouped according to which high-resource language was the source of their subnetwork mask (i.e., which high-resource language is most typologically similar), and we report average typological similarity between transfer and test languages ($\bar\theta$). Lastly, while results are not directly comparable since they are from \emph{zero-shot} testing, we report available UDify (UDF) and UDapter (UDA) scores to give an indication of current state-of-the-art performance; our results are underlined when they outperform these models.}
\label{Table:hm_res}
\end{table*}



\noindent
Yet, when transferring from Norwegian, \meta{}-\snstatic{} and \meta{}-\sndyna{} particularly often underperform compared to \textsc{Meta-Full}, see Table \ref{Table:hm_res}. In contrast, \meta{}-\sndyna{} performs particularly well when transferring from Arabic, similarly \snstatic{} performs especially well when transferring from Czech. Thus, the best approach might be dependent on the relationship between the transfer and test languages, or the properties of the transfer language itself.

We note that despite the observed improvements, overall performance remains low for many languages. Yet we would like to point out that we also find instances where our methods might already make the difference in acquiring a usable system compared to state-of-the-art models. For example, even with few-shot fine-tuning UDify's performance on Faroese OFT only reaches 53.8\% which is much lower than our 70.4\% (\nonep{}-\snstatic), and for Indonesian PUD it reaches 69.0\% versus our 74.9\% (\nonep{}-\snstatic)


\section{Analysis}\label{sec:modelbehaviour}
In this section, we provide more insight into the effects of our methods by analyzing performance with respect to four factors: typological relatedness, data-scarcity, robustness to domain transfer, and ability to predict unseen and rare labels. We focus on the best model from each learning framework: \nonep{}-\snstatic{} and \meta{}-\sndyna{}.

\begin{figure}
    \includegraphics[width=\linewidth]{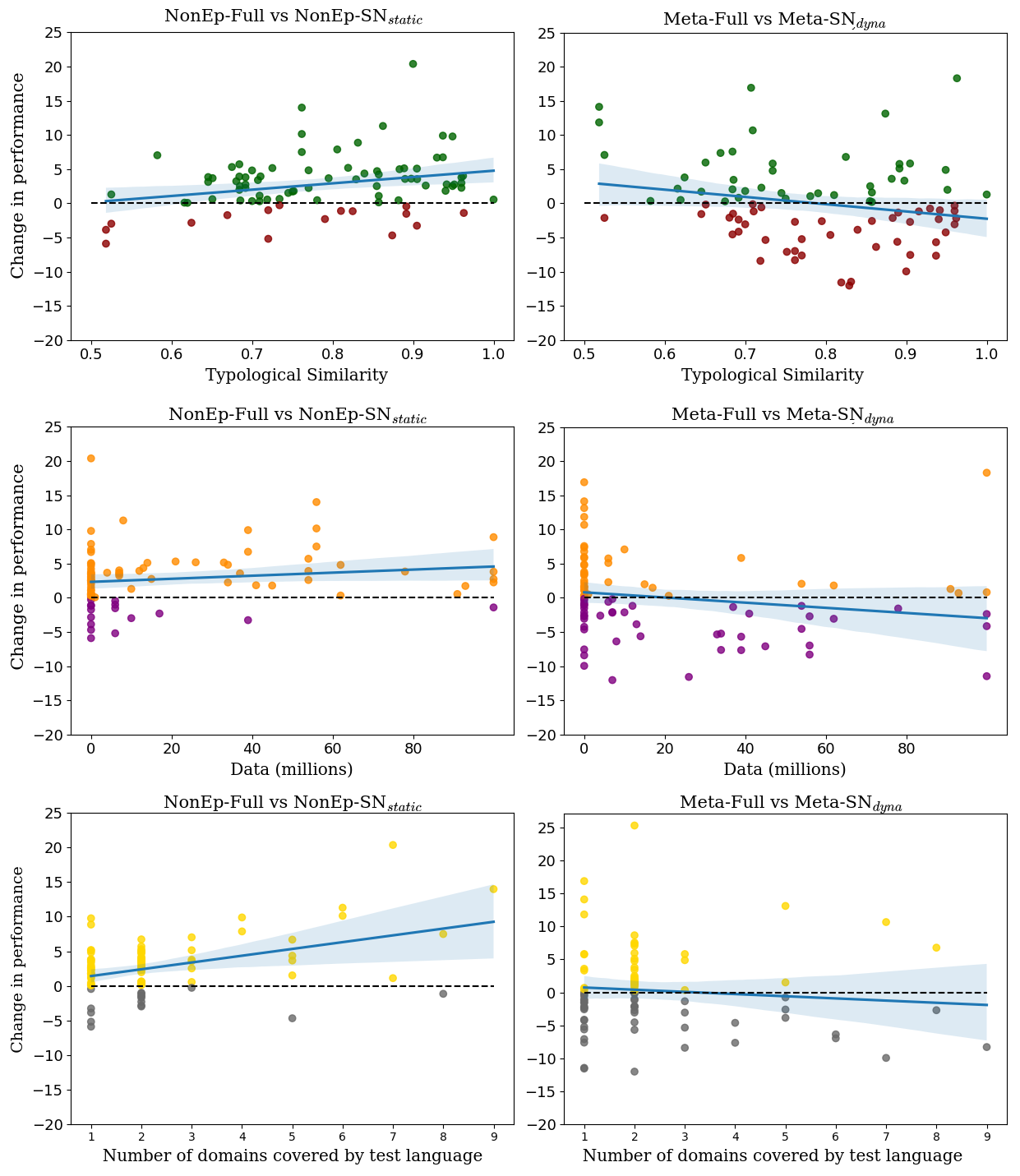}
    \vspace{-0.8cm}
    \caption{Plots of the relationships between a test language's performance gains and:
    (top) how typologically similar the language is to the nearest high-resource fine-tuning language,
    (middle) the amount of in-language data used to pretrain mBERT, and
    (bottom) the number of domain sources represented in its test data.}
    \label{fig:typology-scarcity}
\end{figure}

\paragraph{Typological relatedness}
The languages most similar to a low-resource language are often themselves low-resource, meaning that a low-resource language might be quite dissimilar from all the languages that are resource-rich enough to be used for fine-tuning.
A method that only works well when a very similar high-resource language is available for fine-tuning will not be as useful in practice.
Thus, we want to understand the degree to which our methods depend on similarity to a high-resource fine-tuning language.
In Figure~\ref{fig:typology-scarcity} (top), we plot each test language's performance improvement against its typological closeness to the nearest high-resource fine-tuning, where that distance is as computed using the cosine similarity between the languages' URIEL features. Interestingly, we find that our models show opposite trends: while \nonep{}-\snstatic{} works well for typologically similar languages, the biggest gains from \meta{}-\sndyna{} actually come from less similar languages.

\paragraph{Data scarcity}
Given that language distribution in the mBERT pretraining corpus is very uneven, and 41 of our 74 unique test languages are not covered at all, we want to understand what effect this has on downstream model performance. 
As shown in Figure~\ref{fig:typology-scarcity} (middle), we find that \meta{}-\sndyna{} provides the most benefit to previously unseen languages.
In contrast, more data in pretraining positively correlates with the performance of \nonep{}-\snstatic{}.

\paragraph{Out-of-domain data}
For cross-lingual transfer we often focus on the linguistic properties of source and target languages. However, the similarity of the source and target datasets will also be based on the domains from which they were drawn \citep{glavavs2021climbing}.
For example, our training datasets cover only 11/17 domains, as annotated by the creators of the UD treebank. 
While we acknowledge that it is difficult to neatly separate data based on source domain, we test for a correlation between performance and the proportion of out-of-domain data. Interestingly, we find no clear correlation with the percentage of domains from the test language covered by the transfer language. We do, however, find a strong correlation with the domain diversity of the transfer and test language in general for \nonep{}-\snstatic{}, as shown in Figure~\ref{fig:typology-scarcity} (bottom), where we plot improvements against number of domain sources our test data is coming from (more sources \textrightarrow{} more diversity). In contrast, we see that \meta{}-\sndyna{} remains insensitive to this variable.

\begin{table}[!t]
\centering
    \footnotesize
     \begin{tabular}{
            @{\hspace{0.25\tabcolsep}}
        l
        r@{\hspace{0.2\tabcolsep}}l
            @{\hspace{1.0\tabcolsep}}
        r@{\hspace{0.2\tabcolsep}}l
            @{\hspace{1.0\tabcolsep}}
        r@{\hspace{0.2\tabcolsep}}l
            @{\hspace{0.0\tabcolsep}}
    }
     \toprule
     \multicolumn{1}{r@{}}{\nonep{}-}
     & \multicolumn{2}{c}{\textsc{Full}}
     & \multicolumn{2}{c}{\snstatic{}}
     & \multicolumn{2}{c}{\sndyna{}} \\
    \midrule
     Unseen & 0.04\% &   (3/3)  &  0.003\% &  (1/1)   &  0.004\% &  (2/2)  \\
     Rare   & 12.5\% & (12/50)  &  6.4\%   & (11/41)  &  9.9\%   & (8/49)  \\
    \midrule
    \midrule
    \multicolumn{1}{r@{}}{\meta{}-}
     & \multicolumn{2}{c}{\textsc{Full}}
     & \multicolumn{2}{c}{\snstatic{}}
     & \multicolumn{2}{c}{\sndyna{}} \\
     \hline
     Unseen &    0\% &         &   0\% &        &  \textbf{6.6\%} & (15/23) \\
     Rare   &  3.5\% & (10/39) & 3.0\% & (7/36) & \textbf{21.3\%} & (13/55)\\
    \bottomrule
    \end{tabular}
    \vspace{-2mm}
\caption{Percentages of correctly predicted instances of unseen and rare labels. We also report across how many labels/languages correct predictions were made.}
\label{Table:rareunseen}
\end{table}

\paragraph{Unseen and rare labels}
Lastly, another problem in cross-lingual transfer, especially when fine-tuning on only a few languages, is that the fine-tuning data may not cover the entire space of possible labels from our test data. In principle, only a model that is able to adequately adapt to unseen and rare labels can truly succeed in cross-lingual transfer. Given that we perform few-shot fine-tuning at test time, we could potentially overcome this problem \citep{lauscher2020zero}. Thus, we investigate the extent to which our models succeed in predicting such labels for our test data. We consider a label to be rare when it is covered by our training data, but makes up \textless 0.1\% of training instances (23 such labels). There are 169 unseen labels, thus in total, 192/233 (82\%) of the labels from our test data are rare or unseen during training. 
In Table~\ref{Table:rareunseen}, we report how often each model correctly predicts instances of unseen and rare labels. We find that models differ greatly, and, in particular, \meta{}-\sndyna{} vastly outperforms all other models when it comes to both unseen and rare labels. Upon further inspection, we find that two unseen labels are particularly often predicted correctly: sentence particle (\texttt{discourse:sp}) and inflectional dependency (\texttt{dep:infl}). The former label seems specific to Chinese linguistics and has a wide range of functions e.g.\ modifying the modality of a sentence or its proposition, and expressing discourse and pragmatic information. The latter represents inflectional suffixes for the morpheme-level annotations, something that is unlikely to be observed in morphologically poor languages such as English; but, for instance, Yupik has much of its performance boost due to it.

\begin{table}[t]
    \centering
    \begin{tabular}{lcll}
    \toprule
    Language 
     & \multicolumn{1}{c}{\textsc{Full}}
     & \multicolumn{1}{c}{\snstatic{}}
     & \multicolumn{1}{c}{\sndyna{}} \\
    \midrule
    Arabic    & 68.6 & \textbf{72.9} (13) & 69.1 (28)  \\
    Czech     & 75.4 & \textbf{81.2} (13) & 77.9 (28)  \\
    Estonian  & 65.4 & \textbf{69.2} (37) & 68.3 (28) \\
    Hindi     & 74.4 & \textbf{77.2} (21) & 75.2 (28) \\
    Italian   & 85.0 & \textbf{87.7} (23) & 86.1 (28)  \\
    Norwegian & 73.2 & \textbf{79.8} (24) & 73.6 (28)  \\
    Russian   & 79.5 & \textbf{81.6} (27) & 80.4 (28) \\
    \bottomrule
    \end{tabular}
    \caption{Labeled Attachment Scores for Non-Episodic models on each training language. Number of heads disabled by the subnetwork is shown in parentheses.}
    \label{Table:maskinfo}
\end{table}

\section{Effect of subnetworks at training time }

\subsection{Interaction between subnetworks}
We now further investigate the selected subnetworks and their impact during training. Our findings were similar for meta-learning, so we just focus our analysis here on the non-episodic models. 

Table~\ref{Table:maskinfo} shows how using subnetworks affects performance on the training languages. Training with the subnetworks always improves performance, however, this effect is larger when subnetworks are kept static during training. Moreover, for the static subnetworks, the number of heads that are masked out can vary considerably per language; e.g., for Arabic we only disable 13 heads compared to 37 for Estonian. Yet, we observe similar effects on performance, obtaining $\sim$+4\% improvement for both languages. 
To disentangle how much of the performance gain comes from disabling suboptimal heads vs. protection from negative interference by other languages, we re-train \nonep{}-\snstatic{} in two ways using Czech as a test case: (1) we keep Czech restricted to its subnetwork, but drop subnetwork masking for the other languages, i.e.\ we disable suboptimal heads for Czech, but do not protect it from negative interference; (2) we use subnetworks for all languages \emph{except} Czech, i.e.\ we protect Czech from other languages, but allow it to use the full model capacity.

We find that (1), disabling suboptimal heads only, results in 79.5 LAS on Czech (+4.1\% improvement compared to baseline), while (2), just protection from other languages, results in 80.3 LAS (+4.7\% improvement). This indicates that protection from negative interference has a slightly larger positive effect on the training language in this case. Still, a combination of both, i.e.\ using subnetworks for all fine-tuning languages, results in the best performance (81.2 LAS, a +5.9\% improvement, as reported in Table \ref{Table:maskinfo}). This suggests that the interaction between the subnetworks is a driving factor behind the selective sharing mechanism that resolves language conflicts. We confirm that similar trends were found for the other languages. 

This, however, also means that if the quality of one subnetwork is suboptimal, it is still likely to negatively affect other languages. 
Moreover, analysing the subnetworks can provide insights on language conflicts. For instance, using a subnetwork for only Czech or Arabic results in the biggest performance gains for Norwegian (+7.1\% and +7.3\% compared to the \textsc{Full} baseline), indicating that, in this setup, Norwegian suffers more from interference.

\subsection{Gradient conflicts and similarity}

\begin{table}[!t]
    \begin{tabular}{lcc}
    \toprule
       & Conflicts & Cosine Sim.  \\ 
     \midrule
      \nonep{}-\textsc{Full} &  42\% & 0.03  \\
      \nonep{}-\snstatic{}   &  \textbf{26\%} & 0.05  \\
      \nonep{}-\sndyna{}     & 38\% & \textbf{0.07} \\
     \midrule
       \meta{}-\textsc{Full} & 55\% & $-$0.04 \\
       \meta{}-\snstatic{}   & 54\% &$-$0.02   \\
       \meta{}-\sndyna{}     & \textbf{44\%} & \textbf{0.12} \\
    \midrule
\end{tabular}
\centering
\vspace{-0.2cm}
\caption{We report the percentage of gradient conflicts and average cosine similarity between gradients over the last 50 iterations/episodes for our non-episodic and meta-trained models. We report average results over 4 random seeds.}
\label{Table:grads}
\end{table}

In multilingual learning, we aim to maximize knowledge transfer between languages while minimizing negative transfer between them. In this study, our main goal is to help with the latter. To evaluate the extent to which our methods succeed in doing this, we explicitly test whether we are able to mitigate negative interference by adopting the gradient conflict measure from \citet{yu2020gradient}. They show that \textit{conflicting gradients} between dissimilar tasks, defined as a negative cosine similarity between gradients, is predictive of negative interference in multi-task learning. Similar to \citet{wang2020negative}, we deploy this method in the multilingual setting: we study how often gradient conflicts occur between batches from different languages. 

At the same time, \citet{lee2021sequential} argue that lower cosine similarity between language gradients indicates that the model starts memorizing language-specific knowledge that at some point might cause catastrophic forgetting of the pretrained knowledge. This suggests that, ideally, our approach would find a good balance between minimizing gradient conflicts and maximizing the cosine similarity between the language gradients. 

We quantitatively find that both subnetwork-based methods indeed reduce the percentage of gradient conflicts between languages. Over the last 50 iterations, we find that \nonep{}-\snstatic{} has reduced conflicts by 16\% and \nonep{}-\sndyna{} by 4\% compared to the \nonep{}-\textsc{Full} baseline as reported in Table \ref{Table:grads}. In the meta-learning setup we found an opposite trend where \meta{}-\snstatic{} reduces conflicts by 1\% and \meta{}-\sndyna{} by 11\% over the last 50 iterations compared to \textsc{Meta-Full}. This partly explains why \nonep{}-\snstatic{} and \meta{}-\sndyna{} are found to be the best performing models: they suffer from gradient conflicts the least. 
Interestingly, we do not find that our meta-trained models suffer less from gradient conflicts than the non-episodic models. In fact, while we found that, on average, \textsc{Meta-Full} improves over \nonep{}-\textsc{Full} (recall Table \ref{Table:winning}), its training procedure suffers from 13\% more conflicts, meaning that we do not find meta-learning in itself to be a suitable method for reducing gradient conflicts, but our subnetwork-based methods are.

At the same time, the average cosine similarity between gradients increases when using both subnetwork methods compared to the \textsc{Full} model baselines. We compute the Pearson correlation coefficient between the relative decrease in percentage of gradient conflicts and increase in cosine similarity over training iterations compared to the baselines. We test for statistical significance ($p$-value \textless 0.02), and average results over 4 random seeds. We get statistically significant positive correlation scores of 0.08, 0.16, 0.33 and 0.58 for \nonep{}-\snstatic{}, \nonep{}-\sndyna{}, \meta{}-\snstatic{} and \meta{}-\sndyna{}, respectively. This indicates that our subnetwork-based methods try to minimize negative interference while simultaneously maximizing knowledge transfer.  

\section{Ablations}
To ensure that each of the aspects of our setup are indeed contributing to the improvements shown in our experiments, we retrained models with specific aspects ablated.

\paragraph{Random mask initialization -- Static}
In these experiments, we verify that there is value in using the iterative pruning procedure to generate subnetwork masks (as opposed to the value coming entirely from the mere fact that masks were used).

\begin{figure}[t!]
    \centering
    \includegraphics[width=\linewidth, height=3.5cm]{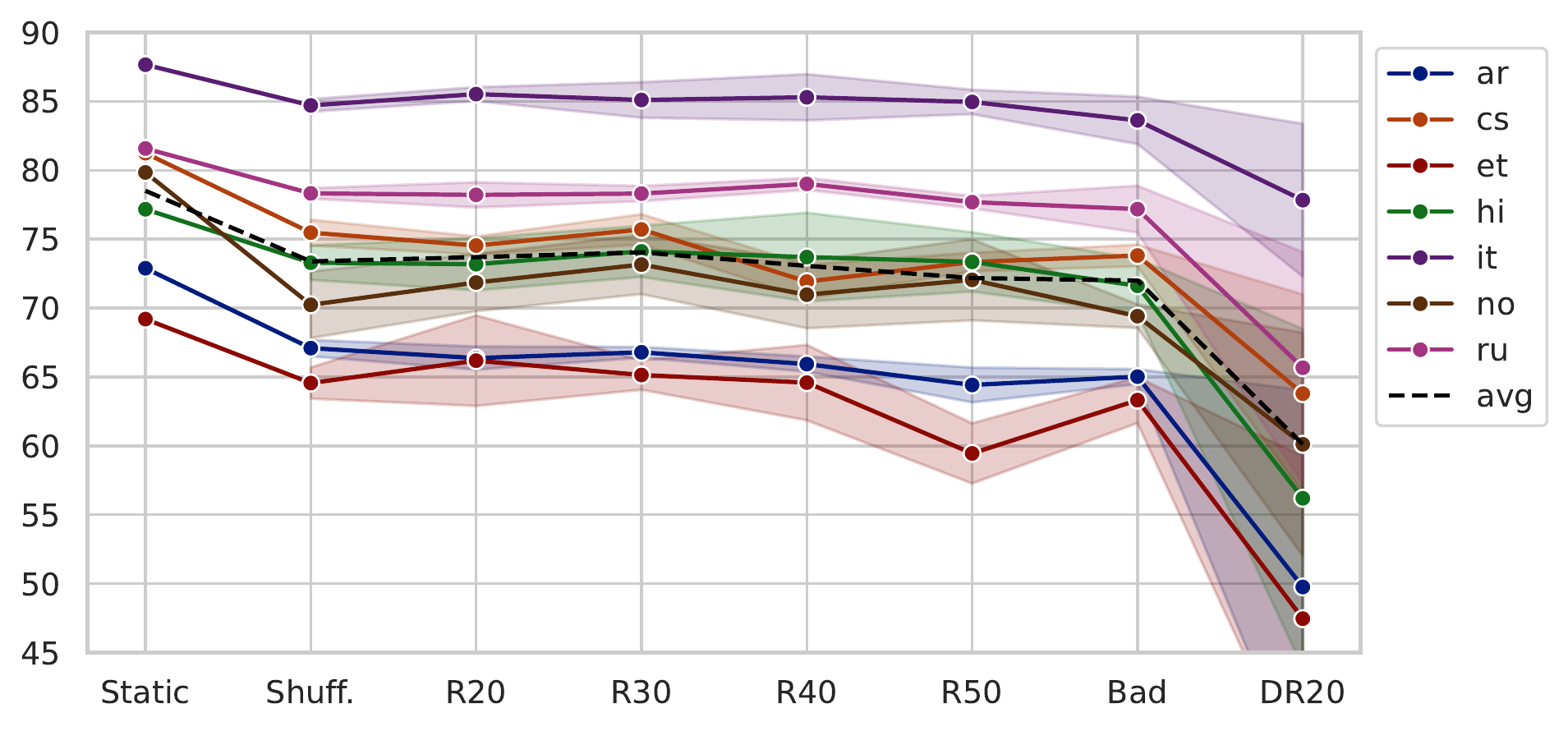}
    \vspace{-0.8cm}
    \caption{Effect of training with masks randomly generated under different constraints (across 3 seeds): shuffled, masking $n$ heads, only select bad heads and start dynamic training from a random subnetwork (DR20).}
    \label{fig:rdmcomp}
\end{figure}

First, we re-trained \nonep{}-\snstatic{}, but swapped out the subnetwork masks derived from iterative pruning with masks containing the same number of enabled heads, but that were randomly generated (Shuffle). 
Second, given that the number of masked heads might be more important than which exact heads are being masked out, we experiment with masking 20, 30, 40, and 50 random heads. 
We find that using the random masks results, on average, in $\sim$5\% performance decreases on the training languages compared to using the subnetworks initialized using importance pruning; see Figure \ref{fig:rdmcomp}. 
In addition, we see that randomly masking out more heads results in further negative effects on performance.


Lastly, given that for many languages our subnetworks mask out very few heads (e.g.\ 13 for Arabic and Czech), we also try swapping these out with ``intentionally bad'' masks, where we randomly choose 20 heads to mask out
, but do not allow any of the heads selected by the real pruning procedure to be chosen (Bad). 
From this, we see that preventing the right heads from being selected for masking 
does result in lower performance versus pure random selection (R20). 

\paragraph{Random mask initialization -- Dynamic}
In these experiments, we verify that there is value in using the iterative pruning procedure to initialize subnetwork masks that will then by dynamically updated during fine-tuning.


We retrained \nonep{}-\sndyna{} 3 times using randomly initialised subnetworks.
Figure \ref{fig:rdmcomp} (DR20) shows that average performance across all test languages drops substantially ($\sim$10\%), making this method considerably worse than any of our other random baselines. We hypothesize that this is because the model is able to correct for any random static subnetwork, but that with dynamic masking, the subnetworks keep changing, which deprives the model of the chance to properly re-structure its information. This also gives us a strong indication that the improvements we observe are not merely an effect of regularization \citep{bartoldson2020generalization}.

\paragraph{Random transfer language} To test the effectiveness of our typology-based approach to selecting which high-resource fine-tuning language's subnetwork should be used for a given test language, we experimented with just picking one of the high-resource languages at random, and found that this performed worse overall, resulting in lower scores for 79/86 test languages.
 
\paragraph{Unstructured pruning}
Our approach relies on the assumption that attention heads function independently. However, attention head interpretability studies have sometimes given mixed results on their function in isolation \citep{prasanna2020bert, clark2019does, htut2019attention}. Moreover, related works commonly focus on unstructured methods \citep{lu2022language, nooralahzadeh2020zero}.
Thus, we compare our strategy of masking whole attention heads against versions of \nonep{}-\snstatic{} and \nonep{}-\sndyna{} that were retrained using subnetwork masks found using the most popular unstructured method, \emph{magnitude pruning}.
In magnitude pruning, instead of disabling entire heads during the iterative pruning procedure, as described in \S\ref{ref:mask-init}, we prune the 10\% of parameters with the lowest magnitude weights across all heads.
Again, we check the development set score in each iteration and keep pruning until reaching \textless95\% of the original performance. Note that we exclude the embedding and MLP layers.\footnote{We recognize that the interaction between the MLPs and attention heads is important, but by focusing on the attention heads, we keep results comparable to importance pruning.}

We find that for both the static and dynamic strategies, 
unstructured pruning performs worse overall, resulting in lower scores for 76\% of test languages, and is especially harmful for dynamic subnetworks (\snstatic{}: 40.4 vs. 39.9, and \sndyna{}: 39.0 vs. 36.7 average LAS). We hypothesize that it might be more difficult to learn to adapt the unstructured masks as there are more weights to learn (weights per head $\times$ heads per layer $\times$ layers).

\section{Conclusion}
We present and compare two methods, i.e. static and dynamic subnetworks, that successfully help us guide selective sharing in multilingual training across two learning frameworks: non-episodic learning and meta-learning. We show that through the use of subnetworks, we can obtain considerable performance gains on cross-lingual transfer to low resource languages compared to full model training baselines for dependency parsing.  Moreover, we quantitatively show that our subnetwork-based methods are able to reduce negative interference. Importantly, our training strategy is data-efficient, requiring vastly less compute time and data compared to current state-of-the-art models. Finally, we extensively analyze the behaviour of our best performing models and show that they possess different strengths, obtaining relatively large improvements on different sets of test languages with often opposing properties. Given that our \meta{}-\sndyna{} model performs particularly well on data-scarce and typologically distant languages from our training languages, this is an interesting approach to further explore in future work on low-resource languages. In particular, it would be interesting to investigate methods to integrate the strengths of \nonep{}-\snstatic{} and \meta{}-\sndyna{} into one model. 

\section*{Acknowledgements}
This project was in part supported by a Google PhD Fellowship for the first author. 
We would like to thank Tim Dozat and Vera Axelrod for their thorough feedback and insights.

\appendix

\section{Training and data details}
 \label{app}
\begin{table}[H]
    \footnotesize
    \begin{tabular}{c|c|c|c|c|c}
    & Family & TB & Train  &Val. & Test \\
    \hline
    ar & Afro-Asiatic & PADT & 6075 & 909 & 680 \\
    cs & Slavic & PDT & 68495  &9270& 10148 \\
    en &  German. &EWT &12543 &2002 & 2077 \\
    hi & Indic & HDTB&  13304 & 1659 & 1684 \\
    it& Roman. & ISDT & 13121 & 564 & 482 \\
    et & Urallic & EDT & 24633 & 3125 & 3214 \\
    no & German. & Norsk & 14174 & 1890 & 1511 \\
    ru & Slavic & SynTag & 48814 & 6584 & 6491 \\
    \end{tabular}
    \caption{Number of sentences in the UD treebanks for our training languages.}
    \label{tab:datasets}
\end{table}
\begin{table}[H]
    \scriptsize
    \begin{tabular}{l|ll}
        & \multicolumn{2}{c}{Inner/Test LR}  \\
        & mBERT & decoder \\
    \hline 
    \nonep{} &  \{\textbf{1e-04}, 5e-05, 1e-05\}  & \{\textbf{1e-03}, 5e-04,        1e-04\}  \\
    Unstructured &  \{1e-04, \textbf{5e-05}, 1e-05\}  & \{1e-03, \textbf{5e-04}, 1e-04\}  \\
    \textsc{Meta-Full} & \{1e-04, 5e-05, \textbf{1e-05}\} &  \{1e-03, \textbf{5e-04}, 1e-04\} \\
    \meta{}-\snstatic{} & \{1e-04, 5e-05, \textbf{1e-05}\} &  \{1e-03, \textbf{5e-04}, 1e-04\} \\
    \meta{}-\sndyna{} & \{\textbf{1e-04}, 5e-05, 1e-05\} &  \{\textbf{1e-03}, 5e-04, 1e-04\} \\
        & \multicolumn{2}{c}{Outer LR}\\
    Meta-All &  \{\textbf{1e-04}, 5e-05, 1e-05 \} & \{\textbf{1e-03}, 5e-04, 1e-04\} \\
\end{tabular}
\centering
\caption{Final selection of learning rates. For all non-episodic models, we use the same learning rates (\nonep{}). Similarly, we found the same optimal hyperparameter values for all outer-loop learning rates of the meta-trained models (Meta-All). Moreover, the hyperparameter selection is performed based on 4 validation languages: Bulgarian, Japanese, Telugu and Persian. }
\label{Table:hp}
\end{table}

All models use the same UDify architecture with the dependency tag and arc dimensions set to 256 and 768 respectively. At fine-tuning stage 1, we train for 60 epochs following the procedure of \citet{langedijk2022meta, kondratyuk201975}. The Adam optimizer is used with the learning rates of the decoder and BERT layers set to 1e-3 and 5e-5 respectively. Weight decay
of 0.01 is applied, and we employ a gradual unfreezing
scheme, freezing the BERT layer weights for the
first epoch. For more details on the training procedure and hyperparameter selection, see \citet{langedijk2022meta}. For fine-tuning on seperate languages to find the subnetworks, we apply the same procedure.

Moreover, we need $\sim$3 hours for pretraining and depending on the training set size $\sim$4 hours per language for fine-tuning and finding a subnetwork (Note that this step is run in parallel for all languages and only needs to be performed once for all models trained with subnetworks). We then only require $\sim$1 hour for non-episodic training or $\sim$6 hours for meta-training. All models are trained on a NVIDIA TITAN RTX. 

\bibliography{tacl2021}
\bibliographystyle{acl_natbib}

\end{document}